\pdfoutput=1

\documentclass[11pt]{article}

\usepackage[]{acl}

\usepackage{times}
\usepackage{latexsym}

\usepackage{graphicx}
\usepackage{xcolor}
\usepackage{multirow}

\usepackage{amsmath}
\usepackage{amssymb}

\usepackage[T1]{fontenc}

\usepackage[utf8]{inputenc}

\usepackage{microtype}

\title{
\vspace*{-0.5in}
{{\small \hfill ACL'2022 findings}\\
\vspace*{.25in}} 
Relevant CommonSense Subgraphs for "What if..." Procedural Reasoning}

 \author{Chen Zheng \\
  Michigan State University  \\
  \texttt{zhengc12@msu.edu} \\\And
  Parisa Kordjamshidi \\
  Michigan State University \\
  \texttt{kordjams@msu.edu} \\}

\begin{document}
\maketitle
\begin{abstract}
We study the challenge of learning causal reasoning over procedural text to answer "What if..." questions when external commonsense knowledge is required. We propose a novel multi-hop graph reasoning model to 1) efficiently extract a commonsense subgraph with the most relevant information from a large knowledge graph; 2) predict the causal answer by reasoning over the representations obtained from the commonsense subgraph and the contextual interactions between the questions and context. We evaluate our model on WIQA benchmark and achieve state-of-the-art performance compared to the recent models.

\end{abstract}

\section{Introduction}
\label{sec:intro}

\begin{figure}
\centering
\includegraphics[width=0.40\textwidth,height=150pt]{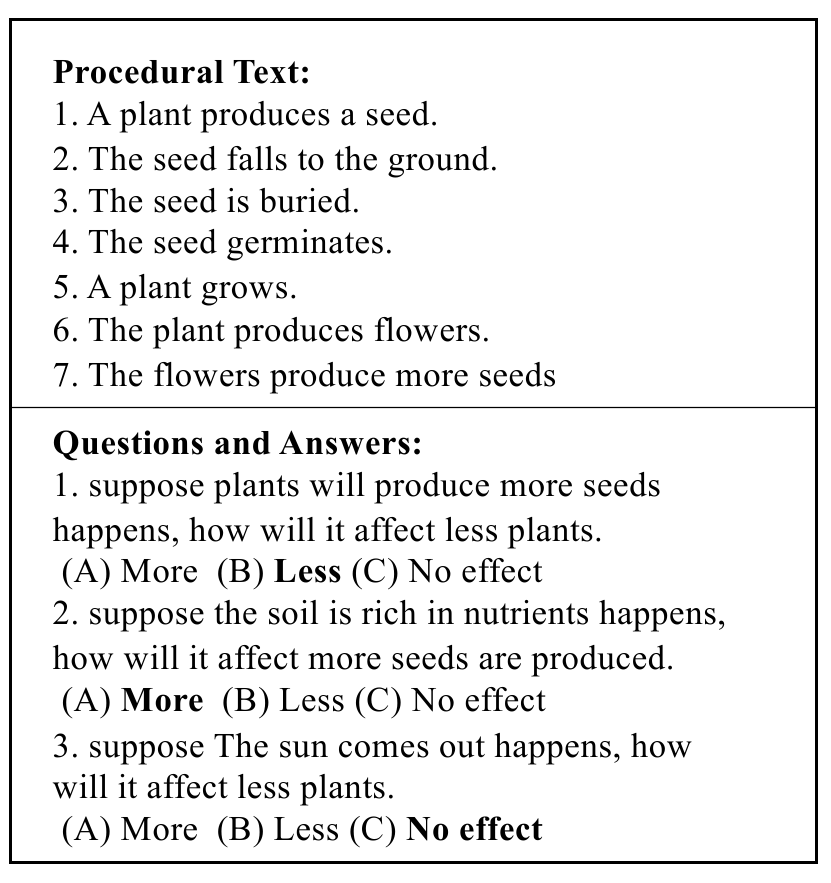}
\caption{WIQA contains procedural text, and different types of questions. The bold choices are the answers. }
\label{fig:wiqa_example}
\vspace{-6mm}
\end{figure}

In recent years, large-scale pre-trained language models (LMs) have made a breakthrough progress and demonstrate a high performance in many NLP tasks, including procedural text reasoning ~\cite{Tandon2019WIQAAD,rajagopal-etal-2020-ask}.
There is a large amount of knowledge that is stored implicitly in language models that help in solving various NLP tasks~\cite{devlin-etal-2019-bert}.
When we reason over text, sometimes, the knowledge contained in a given text is sufficient to predict the answer, as it is shown in the question $1$ of Figure~\ref{fig:wiqa_example}. 
This knowledge is directly encoded and used by LMs models~\cite{Tandon2019WIQAAD}.
However, there are many cases in which the required knowledge is not included in the procedural text itself. 
For example, for the question $2$ in Figure~\ref{fig:wiqa_example}, the information about the ``nutrient'' on the seeds does not exist in the procedural text. Therefore, 
the external commonsense knowledge is required. 

\begin{figure*}
\centering
\includegraphics[width=0.99\textwidth,height=95pt]{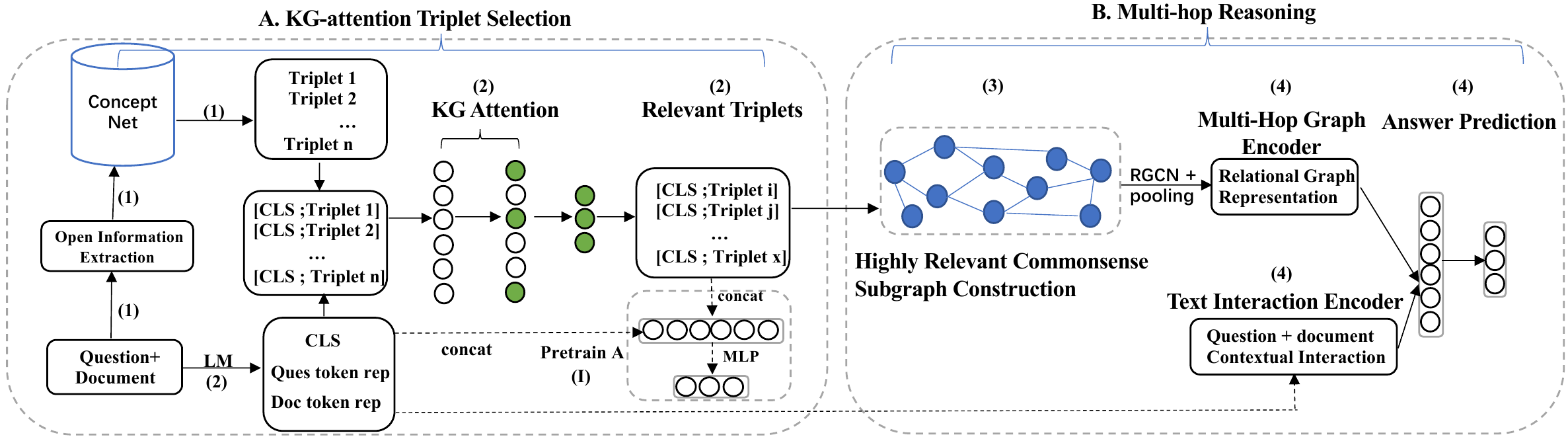}
\caption{MRRG Model is composed of Candidate Triplet Extraction, KG Attention, Commonsense Subgraph Construction, Text encoder with contextual interaction, Graph Reasoning, and Answer prediction modules.\label{fig:architecture}}
\vspace{-4mm}
\end{figure*}


There are several existing resources that contain world knowledge and commonsense. Examples are knowledge graphs (KGs) like ConceptNet~\cite{speer2017conceptnet} and ATOMIC~\cite{sap2019atomic}. 
Looking back at the question 2, 
we observe that through providing the external knowledge triplets (nutrient, relatedto, soil) and (soil, relatedto, seed) derived from ConceptNet, we can build an explicit reasoning chain and choose an explainable answer.




Two challenges exist in procedural text reasoning and using external KBs. The first challenge is effectively extracting the most relevant external information and reducing the noise from the KB.
The second challenge is reasoning over the extracted knowledge.
Several works enhance the QA model with commonsense knowledge~\cite{lin-etal-2019-kagnet,lv2020graph}. 
However, the noisy knowledge from KG will seriously mislead the QA model in predicting the answer. Moreover, using KBs is often investigated in the tasks that perform QA directly over KB itself, such as CommonsenseQA~\cite{talmor-etal-2019-commonsenseqa}, etc. There are less sophisticated techniques proposed for using external knowledge explicitly (i.e. not through training LMs) in reading comprehension for aiding QA over text.
REM-Net~\cite{huang2020rem} is the only work that uses commonsense for WIQA and uses a memory network to extract the external triplets to solve the first challenge. However, this work has no reasoning process over the extracted knowledge and uses a simple multi-head attention operator to predict the answer. EIGEN~\cite{madaan2020eigen} constructs an influence graph to find the chain of reasoning given procedural text. However, EIGEN cannot deal with the challenge when the required knowledge is not in the given document. 

To solve these two challenges, we propose a \textbf{M}ulti-hop \textbf{R}easoning network over \textbf{R}elevant CommonSense Sub\textbf{G}raphs (MRRG) for casual reasoning over procedural Text.
Our motivation is to effectively and efficiently extract the most relevant information from a large KG to help procedural reasoning.
First, we extract the entities, retrieve related external triplets from KG, and learn to extract the most relevant triplets to a given the procedure and question input by a novel KG attention mechanism. Then, we construct a commonsense subgraph based on the extracted KG triplets in a pipeline.  We use the extracted subgraphs as a part of end-to-end QA model to help in filling the knowledge gaps in the procedure and performing multi-hop reasoning.
The final model predicts the causal answer by reasoning over the contextual interaction representations over the question and the document and learning graph representations over the KB subgraphs. We evaluate our MRRG on the ``what if'' WIQA benchmark. MRRG model achieves SOTA and brings significant improvements compared to the existing baselines.

The contributions of our work are: {\bf 1)} We train a separate module that extracts the relevant parts of the KB given the procedure and question to avoid the noisy and inefficient usage of the information in large KBs. 2) We design an end-to-end model that uses the extracted QA-dependent KB as a subgraph to guide the reasoning over the procedural text to answer the questions. 3) Our MRRG achieves SOTA on the WIQA benchmark.

\section{Model Description}
\label{sec:model}

\subsection{Problem Formulation and Overview}
Formally, the problem is to predict an answer $a$ from a set of pre-defined answers given input question $q$, a document $\mathcal{C}$ which is composed of several sentences $\mathcal{C} = \{s_1, \dots, s_n\}$, and a large knowledge graph KG.


Figure~\ref{fig:architecture} shows the proposed architecture. 
(1) We extract the entities from question and context in preprocessing step and use them to retrieve the set of \textbf{candidate triples} from the ConceptNet. 
(2) We train the \textbf{KG Attention} module to extract the most relevant triplets given the procedure and question and reduce the noisy concepts from candidate triplets.
(3) We augment the \textbf{commonsense subgraph} based on the relevant triplets.
(4) We train a model that uses two components, the commonsense subgraph as a relational graph network and a text encoder including question and document to do \textbf{procedural reasoning}.
Below, we describe the details of each module.

\subsection{Candidate Triplet Extraction from KG}
\label{sec:triplet_extraction}
Given the input $q$ and $\mathcal{C}$, we extract the contextual entities (concepts) by a open Information Extraction (OpenIE) model~\cite{stanovsky-etal-2018-supervised}. 
For each extracted entity $t_{in}$, we retrieve the relational triplets
$ t = (t_{in}, r, t_{out})$ from KG, where $t_{out}$ is the concept taken from ConceptNet and $r$ is a semantic relation type. 
We then apply a pre-trained Language Model, RoBERTa, to obtain the representation of each triplet: 
$E^{t} = f_{LM}([t_{in}, r, t_{out}]) \in \mathbb{R}^{3 \times d},$
where $f_{LM}$ denotes the language model operation and the triplets are given as a sequence of concepts and relations to the LM.

\subsection{KG Attention}
\label{sec:kg_attention}
\begin{figure}[ht!]
\centering
\includegraphics[width=0.48\textwidth,height=85pt]{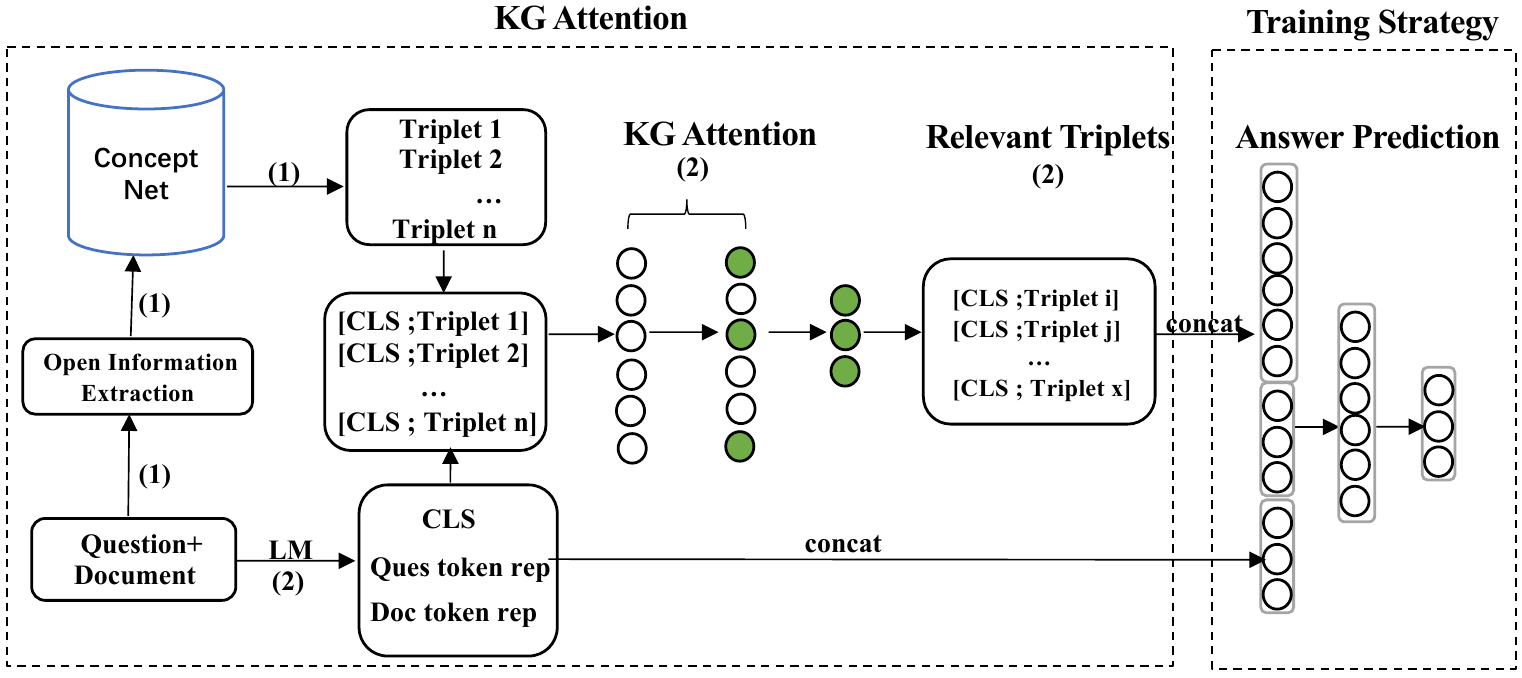}
\caption{The architecture of training the KG Attention module.}
\label{fig:pretrain}
\vspace{-4mm}
\end{figure}

The KG attention module is shown in Figure~\ref{fig:architecture}-A and Figure~\ref{fig:pretrain}. 
We concatenate $q$ and $\mathcal{C}$ to form $Q = [[CLS];q;[SEP]; \mathcal{C}]$,
where [CLS] and [SEP] are special tokens in the LMs tokenizer process~\cite{liu2019roberta}.
We use RoBERTa to obtain the list of token representations $E_{[CLS]}$,  $E_{q}$, and $E_\mathcal{C}$. $E_{[CLS]}$ is the summary representation of the question and paragraph, $E_{q}$ is the list of the question tokens embeddings, and $E_\mathcal{C}$ is the list of the paragraph tokens embeddings output of Roberta.

Given triplet $E^{t}$ that is generated based on the triplet extraction described in Section \ref{sec:triplet_extraction}, we build a context-triplet pair $E_{z}^{t} = [E_{[CLS]};E^{t}_{in};E^{t}_{r};E^{t}_{out}]$, 
where $E^{t}_{in}$ is the representation of the head entity from text, $E^{t}_{out}$ is the representation of the tail entity from KG, and $E^{t}_{r}$ is the representation of the relation.
Afterwards, we compute context-triplet pair attention and a softmax layer to output the \textbf{C}ontext-\textbf{T}riplet pairwise importance \textbf{S}core 
$CTS$. 
The process is computed as follows:
$CTS_t = \frac{ \exp \left( MLP(E_{z}^{t}) \right) } {\sum_{j=1}^m \exp \left( MLP(E_{z}^{t}) \right) }.$

Then we choose the top-$k$ relevant triplets with the top $CTS$ scores and then 
use the relevant triplets to construct the subgraph. For each selected triplet, we obtain the triplet representation 
$E'^{t} = [E'^{t}_{in}, E^{t}_{r}, E'^{t}_{out}] \in \mathbb{R}^{3 \times d}$, where $E'^{t}_{in} = f_{in}([CTS_t \cdot E^{t}_{in}; CTS_t \cdot E^{t}_{r}])$ and $E'^{t}_{out} = f_{out}([CTS_t \cdot E^{t}_{out}; CTS_t \cdot E^{t}_{r}])$. Notice that $f_{in}$ and $f_{out}$ are MLP layers, $[;]$ is the concatenation, and $[\cdot]$ is the 
scalar product.

\subsection{Commonsense Subgraph Construction}

We construct the subgraph $G_{s}$ based on the relevant triplets from KG attention for each question and answer pair. 
We add more edges to the subgraph as follows:
Two entities in the triplets will have an edge if a relation $r$ in the KG exists between them.
The assumption is that the augmented commonsense subgraph will contain the reasoning paths. We use $E'^{t}_{in}$ and $E'^{t}_{out}$ for the KG subgraph initial node representation $h^{(0)}$ which is used in RGCN formulation in Section~\ref{sec:graph_learning}.

\subsection{Procedural Reasoning}
\label{sec:graph_learning}
Procedural Reasoning composes of two parts: Multi-Hop Graph Reasoning and Text Contextual Interaction Encoder. \\
\noindent \textit{(I) Multi-Hop Graph Reasoning:} this is the Graph Reasoning part of Figure~\ref{fig:architecture}-B. Given the subgraph $G_s$,
we use RGCN~\cite{schlichtkrull2018modeling} to learn the representations of the relational graph. RGCN learns graph representations by aggregating messages from its direct neighbors and relational semantic edges. 
The $(l+1)$-th layer node representation $h_i^{(l+1)}$ is updated based on the neighborhood node representations $h_j^l$ from the $l$-layer multiplied by the relational matrices 
$W_{r_1}^{(l)},\dots,W_{r_{|R|}}^{(l)}$. 
The representation $h_i^{(l+1)}$ is computed as follows: 
$$h_i^{(l+1)}= \sigma ( \sum_{r \in \mathcal{R}}\sum_{j \in N^r_i} \frac{1}{|N^r_i|}W_r^{(l)} h_j^{(l)} + W_0^{(l)}h_i^{(l)} ),$$
where $\sigma$ denotes a non-linear activation function,
$N^r_i$ represents a set that includes neighbor indices of node $i$ under semantic relation $r$. Finally, we obtain the $E_{G_s}$ after several hops of message passing.

\noindent \textit{{(II) Text Contextual Interaction Encoder:} }
We have obtained the contextual token representations 
$E_{[CLS]}$,  $E_{q}$, and $E_\mathcal{C}$ 
in the KG attention module that described in Section \ref{sec:kg_attention}. 
Followed by~\citeauthor{seo2016bidirectional}, we utilize Bi-DAF style contextual interaction module to feed $E_{q}$ and $E_\mathcal{C}$ to Context-to-Question Attention $E_{\mathcal{C} \rightarrow q} = softmax(sim(E_q^T, E_\mathcal{C}))E_q $ and Question-to-Context Attention $E_{q \rightarrow \mathcal{C}}$ to obtain the contextual interaction between question and context. 
Then we use LSTM to obtain the hidden state representations:
$F_{q \rightarrow \mathcal{C}} = LSTM(E_{q \rightarrow \mathcal{C}})$, and  
$F_{\mathcal{C} \rightarrow q} = LSTM(E_{\mathcal{C} \rightarrow q}).$

\subsection{Answer Prediction}
We concatenate $E_{[CLS]}$,  $F_{q \rightarrow \mathcal{C}}$, $F_{\mathcal{C} \rightarrow q}$, and 
the compact subgraph representation $E^{'}_{G_s}$ obtained from attentive pooling,
and use it as the final representation:
$ F = [E_{[CLS]}; F_{q \rightarrow \mathcal{C}}; F_{\mathcal{C} \rightarrow q}; E^{'}_{G_s}]. $
Then we utilize a classifier $\mbox{MLP}\left( F \right)$ to predict the answer.
Our MRRG has two separate training modules used in a pipeline for triplet selection and procedural reasoning. 

\noindent \textit{ (I) Training KG Attention for Triplet Selection:} 
Figure~\ref{fig:pretrain} and the left block of Figure~\ref{fig:architecture} show the same triplet selection model. The architecture 
of Figure~\ref{fig:architecture}.B is taken and 3 extra MLP layers added to it for training as shown in Figure~\ref{fig:pretrain}. 
The MLP is applied on the concatenation of 
the concatenation of $[E_{[CLS]};E_q;E_\mathcal{C};E'^{t}_1;\dots;E'^{t}_k]$ to predict the answer.
We use the cross-entropy as the loss function to train the model. 

\noindent \textit{ (II) Training End-to-End MRRG:}
After pre-training the KG attention, we keep the learned parameters and extract the most relevant concepts and construct the multi-relational commonsense subgraph $G_s$. 
We combine subgraph representation and text interaction representation as input to train the answer prediction module by cross-entropy loss. 


\section{Experiments and Results}

We implemented our MRRG framework using PyTorch~\footnote{Our code is available at \url{https://github.com/HLR/MRRG}.}.
We use a pre-trained RoBERTa~\cite{liu2019roberta} to encode the contextual information in the input. 
The maximum number of triplets is $50$ and 
the maximum number of nodes in the graph is $100$. Further details of hyper-parameters of the graph are shown in Table~\ref{tab:ablation}.
The maximum number of words for the paragraph context is $256$. For the graph construction module, we utilize \textit{open Information Extraction} model~\cite{stanovsky-etal-2018-supervised} from AllenNLP\footnote{\url{https://demo.allennlp.org/open-information-extraction.}} to extract the entities. 
The maximum number of hops for the graph module is $3$.
The learning rate is $1e-5$. The model is optimized using Adam optimizer~\cite{Kingma2015AdamAM}.

\subsection{Datasets}
WIQA is a large dataset for ``what if'' causal reasoning.
WIQA contains three types of questions: 1) the questions can be directly answered based on the text, called in-paragraph questions. 2) the questions require external knowledge to be answered, called out-of-paragraph questions, and 3) irrelevant causes and effects, called no-effect questions. WIQA contains 29808 training samples, 6894 development samples, 3993 test samples (test V1), and 3003 test samples (test V2). 

\subsection{Baseline Description}
\label{sec:baseline}

We briefly describe the most recent baselines that use the Transformer-based language model as the backbone. We separately fine-tune the BERT and RoBERTa as the first two baselines.

\noindent \textbf{EIGEN}~\cite{madaan2020eigen} is a baseline that builds an event influence graph based on a document and leverages LMs to create the chain of reasoning to predict the answer. However, EIGEN does not use any external knowledge to solve the problem.

\noindent \textbf{Logic-Guided}~\cite{asai2020logic} is a baseline that combines neural networks and logic rules.
Specifically, the Logic-Guided model uses logic rules including symmetry and transitivity rules to augment the training data. Moreover, the base language model uses the rules as a regularization term during training to impose the consistency between the answers of multiple questions.  

\noindent \textbf{RGN}~\cite{ijcai2021-553} is the recent SOTA baseline that utilizes a gating network~\cite{zheng-etal-2020-cross} to effectively filter out the key entities and relationships in the given document and learns the contextual representations to predict the answer. RGN does not consider the external knowledge for procedural reasoning challenges. 

\noindent \textbf{REM-Net}~\cite{huang2020rem} proposes a recursive erasure memory network to find out the causal evidence. Specifically, REM-Net refines the evidence by a recursive memory mechanism and then uses a generative model to predict the causal answer. REM-Net is the only work that uses external knowledge for WIQA. REM-Net uses the external knowledge by training an attention mechanism that considers the KG triplet representations for finding the answer. It does not explicitly select the most relevant triplets as we do, and the graph reasoning is not exploited for finding the chain of reasoning.


\begin{table}[ht!]
\begin{center} \small
\resizebox{0.49\textwidth}{15mm}{
\begin{tabular}{|l|ccc|c|}
\hline
 Models            & in-para  & out-of-para & no-effect & Test V1 Acc \\ 
\hline
\emph{Majority}      &45.46 &49.47 & 55.0 &30.66 \\
\emph{Polarity}         &76.31 &53.59 & 27.0 &39.43 \\
\emph{Adaboost}~\cite{Freund1995ADG}        & 49.41 & 36.61  &48.42  &43.93   \\
emph{Decomp-Attn}~\cite{Parikh2016ADA}      &56.31 &48.56 &73.42 & 59.48\\
\hline
\emph{BERT (no para)}~\cite{Devlin2019BERTPO}     &60.32 &43.74 &84.18 &62.41 \\
\emph{BERT}~\cite{Tandon2019WIQAAD}     &79.68 & 56.13 & 89.38 & 73.80 \\
\emph{EIGEN}~\cite{madaan2020eigen}     & 73.58 & 64.04 & 90.84 & 76.92 \\

\emph{REM-Net}~\cite{huang2020rem}     & 75.67 & 67.98 & 87.65 & 77.56 \\
\emph{Logic-Guided}~\cite{asai2020logic}     & - & - & - & 78.50 \\
\emph{RoBERTa+KG-attention Triplet Selection}     &72.21 &   64.60 & 89.13 & 75.22 \\
\textbf{\emph{MRRG} (RoBERTa-base)} & \textbf{79.85} & \textbf{69.93} &\textbf{91.02} &\textbf{80.06} \\
\hline
Human     & - & - & - & 96.33\\
\hline
\end{tabular}
}
\end{center}
\caption{Model Comparisons on WIQA test V1 dataset.}

\label{table:results_wiqa_main_1}
\vspace{-5mm}
\end{table}



\begin{figure*}
  \centering
  \includegraphics[width=0.73\textwidth,height=95pt]{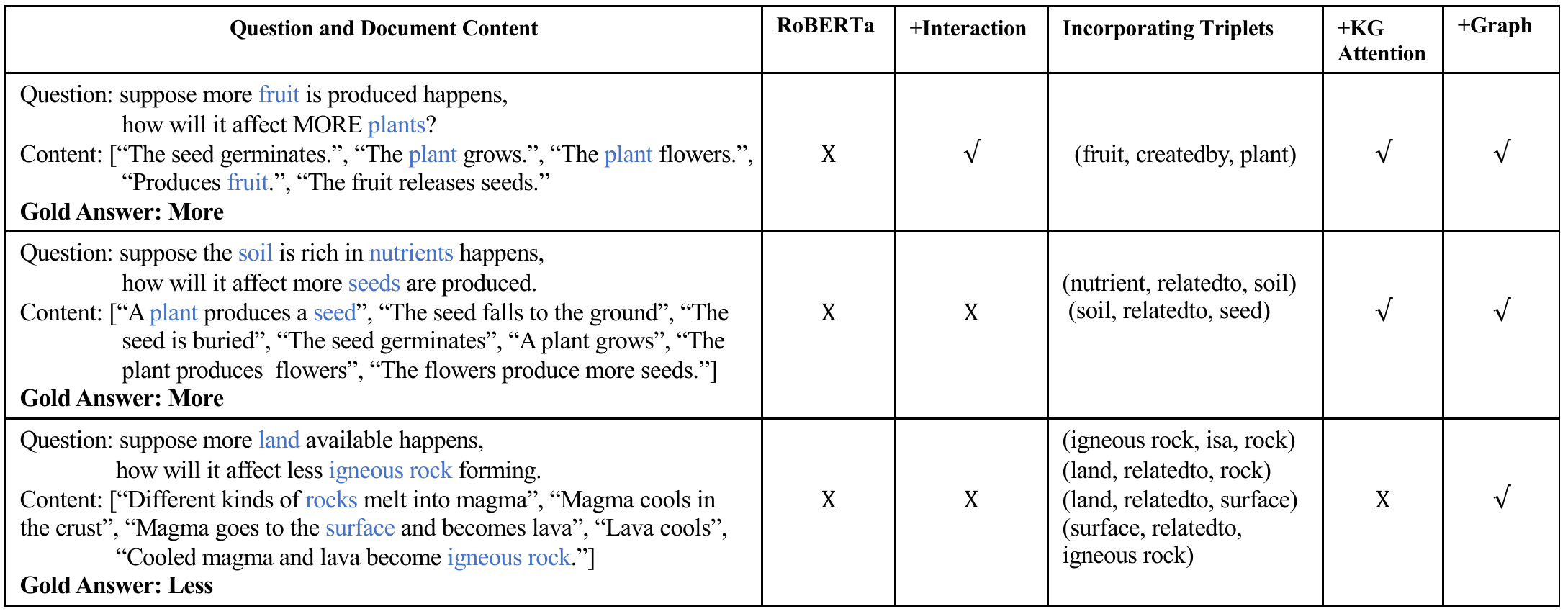}
  \includegraphics[width=0.26\textwidth, height=80pt]{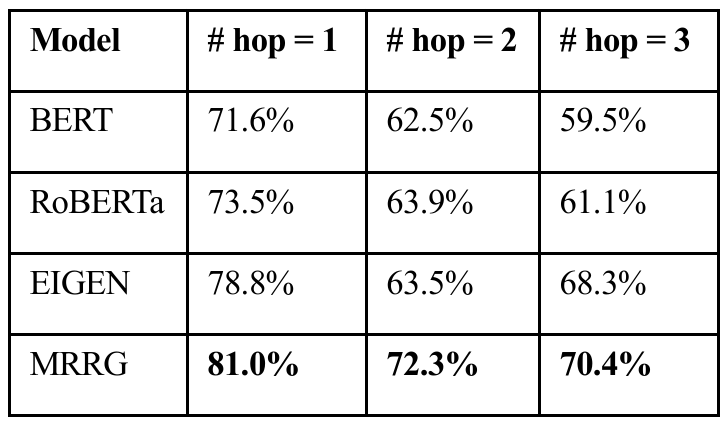}
  \caption{Left: Case study of the MRRG Framework. ``+interaction'' means adding the contextual interaction module. ``KG ATTN'' means adding the KG Attention Triplet Selection module.  'X' indicates the model failed to predict the correct answer and ``$\checkmark$'' means the prediction was successful with the included module. Right: Comparing the results over different number of hops.}
  \label{fig:case_study}
  \vspace{-3mm}
\end{figure*}

\subsection{Results}
\label{sec:result}

Table~\ref{table:results_wiqa_main_1} and Table~\ref{table:results_wiqa_main_2} show the performance of MRRG on the WIQA task compared to other baselines 
on two different test sets V1 and V2.
First,
Both tables show that
our proposed KG Attention triplet selection model outperforms the RoBERTa and has $3.3\%$ improvement on the out-of-para category.
Second, our MRRG achieves SOTA results compared to all baseline models. 
MRRG achieves the SOTA on both in-para, out-of-para, and no-effect questions in WIQA V1 and V2.

\begin{table}[ht!]
\begin{center} \small
\resizebox{0.49\textwidth}{15mm}{
\begin{tabular}{|l|ccc|c|}
\hline
 Models            & in-para  & out-of-para & no-effect & Test v2 Acc \\ 
\hline
  \emph{Random}      &33.33& 33.33& 33.33& 33.33 \\
  \emph{Majority}      & 00.00 & 00.00 & 100.0 & 41.80 \\
\hline
  \emph{BERT}     & 70.57 & 58.54 & 91.08 & 74.26\\
  \emph{REM-Net}    & 70.94 & 63.22 & 91.24 & 76.29 \\
  REM-Net (RoBERTa-large) & 76.23 & 69.13 & 92.35 & 80.09 \\
  \emph{QUARTET (RoBERTa-large)} & 74.49 & 65.65 & 95.30& 82.07 \\
  \cite{rajagopal-etal-2020-ask}    & & & & \\  
  \emph{RGN~\cite{ijcai2021-553}} &75.91 &   66.15 & 92.12 & 79.95 \\
  \emph{RoBERTa+KG Attention Triplet Selection}     &70.02 &   62.30 & 91.23 & 75.86 \\
\textbf{\emph{MRRG} (RoBERTa-base)} & 76.80 & 67.83 & 92.28 & 80.39 \\
   \textbf{\emph{MRRG (RoBERTa-large)}} & \textbf{78.82} &\textbf{71.10} & 93.53 & \textbf{82.95} \\
\hline
Human     & - & - & - & 96.30\\
\hline
\end{tabular}
}
\end{center}
\caption{Model Comparisons on WIQA test V2 dataset.}

\label{table:results_wiqa_main_2}
\vspace{-4mm}
\end{table}

\section{Analysis}

\subsection{Effects of Using External Knowledge}
\label{sec:ana_kb}
In the WIQA, all the baseline models achieve significantly lower accuracy in the out-of-para than in-para and no-effect categories. 
MRRG achieves SOTA in the out-of-para category because of using the highly relevant commonsense subgraphs and the combination of reasoning over text interaction and the graph reasoning modules. As is shown in table~\ref{table:results_wiqa_main_2}, the advantage of the MRRG model is reflected on out-of-para questions. MRRG improves $4.61\%$ over REM-Net.
Notice that REM-Net is the only model that utilizes external knowledge on WIQA.
Figure~\ref{fig:case_study} shows a case in which the ``soil'' and ``nutrient'' only appear in the question and do not exist in the text. 
The baseline models fail to answer this out-of-para question due to missing external knowledge. 
However, our model predicts the correct answer by explicitly incorporating the (nutrient, relatedto, soil), (soil, relatedto, seed) that connects the critical information between the question and document.


\begin{table}[ht!]
    \begin{center}
    \small
    \resizebox{0.40\textwidth}{12mm}{
        \begin{tabular}{|c|c|c|}
            \hline
            Ablation & Model & Dev Acc\\
            \hline
            Text only & RoBERTa-base & 75.51\% \\
            \hline
            Text only & KG Attention Triplet Selection & 77.39\%\\
            \hline
                    & GNN dim=50	 &   79.18\% \\
            Text+Graph & GNN dim=100 &	80.30\%  \\
                    & GNN dim=200 &	79.88\%  \\
            \hline
        \end{tabular}
    }
    \end{center}
    \caption{Ablation and hyper-para. choices on WIQA. ``GNN dim'' is the dimension of graph representation.}
    \label{tab:ablation}
    \vspace{-4mm}
\end{table}

\subsection{Relational Reasoning and Multi-Hops}
\label{sec:ana_hop}
Both in-para and out-of-para question types require multiple hops of reasoning to find the answer in the WIQA.
As shown in the right side of Figure~\ref{fig:case_study}, the MRRG model accuracy improved $2\%$ for $1$ hop, $8\%$ for $2$ hops, and $2\%$ for $3$ hops compared to EIGEN.
MRRG made a sharp improvement in reasoning with multiple hops due to the relational graph reasoning and the effectiveness of the extracted commonsense subgraph. 
We study some cases to analyze the multi-hop reasoning and the reasoning chains.
In the third case in Figure~\ref{fig:case_study},
the extracted relevant triplets (land, relatedto, surface), (surface, relatedto, igneous rock) construct a two-hop reasoning chain ``land$\rightarrow$surface$\rightarrow$igneous rock'' that helps MRRG to find the correct answer. 






\subsection{Ablation Study}
Table~\ref{tab:ablation} shows the ablation study results of MRRG using WIQA. Firstly, we remove the commonsense subgraph and graph network. The accuracy decreases $3.4\%$ compared to MRRG. Second, we remove the contextual interaction module and the accuracy decreases $1.3\%$.
In an additional experiment, we use the KG attention triplet selection module to directly predict the answer without the pipeline of constructing the subgraph and using the graph reasoning module. We show the result as KG Attention Triplet Selection in Table \ref{tab:ablation}. The result shows that removing the triplet selection module decreases the accuracy by $1.8\%$. 
In the same table~\ref{tab:ablation}, we report results about the impact of including the relation types in the RGCN graph and the influence of changing the dimensionality of the node representations in the model.

\section{Conclusion} 
We propose MRRG model for using external knowledge graph in reasoning over procedural text. 
Our model extracts a relevant subgraph for each question from the KG and uses that knowledge subgraph for answering the question.
The extracted subgraph includes the reasoning path for answering the question and helps multi-hop reasoning to predict an explainable answer.
We evaluate MRRG on the WIQA and achieve SOTA performance.

\section*{Acknowledgments}
We thank all reviewers for their suggestions and helpful comments. This project is supported by National Science Foundation (NSF) CAREER award $\#$2028626 and partially supported by the Office of Naval Research (ONR) grant  $\#$N00014-20-1-2005.


\bibliography{acl2022}
\bibliographystyle{acl_natbib}

\end{document}



\appendix
\section{Appendix}
\label{sec:appendix}

\subsection{Implementation Details}

We implemented our MRRG framework using PyTorch~\footnote{Our code will be available if the paper is accepted.}.
We use a pre-trained RoBERTa~\cite{liu2019roberta} to encode the contextual information in the input. 
The maximum number of nodes in the graph is 50. 
The maximum number of words for the paragraph context is 256. For the graph construction module, we utilize a deep BiLSTM open Information Extraction model~\cite{stanovsky-etal-2018-supervised} from AllenNLP\footnote{\url{https://demo.allennlp.org/open-information-extraction.}} to extract the entities. The learning rate is $1e-5$. The model is optimized using Adam optimizer~\cite{Kingma2015AdamAM}.

\subsection{Multi-Hop Reasoning Models}

Various Multi-Hop Reasoning Models have been proposed~\cite{Tu2019SelectAA, Xiao2019DynamicallyFG, fang-etal-2020-hierarchical, zheng-kordjamshidi-2020-srlgrn, madaan2020eigen} that utilize graph neural models (GNN). These GNN models include GAT~\cite{Velickovic2018GraphAN}, graph recurrent network~\cite{Song2018AGM}, and GCN~\cite{kipf2017semi}. GNN models update node embeddings by aggregating messages from their direct neighbors. However, these models are proposed for non-relational graphs, which ignore the semantic relationships. Moreover, the graphs do not consider the external knowledge that does not appear in the text. Unlike these works, our model constructs the semantic relationships as the edge in the subgraph and considers the external knowledge. We use RGCN~\cite{schlichtkrull2018modeling} to learn the relational graph representations with the relational multi-hop message passing. 

\subsection{External Knowledge}

Some research works try to utilize external knowledge to help in reasoning QA tasks~\cite{huang2020rem,lin-etal-2019-kagnet,feng-etal-2020-scalable}. These works extract the knowledge from large knowledge resources which were deliberately created to capture the commonsense information. This includes ConceptNet~\cite{speer2017conceptnet}, Cyc~\cite{lenat1995cyc}, and ATOMIC~\cite{sap2019atomic}. 
\citeauthor{huang2020rem} incorporates commonsense knowledge into a memory network, called Rem-Net on type1 QA. To our knowledge, this is the only work that uses commonsense for WIQA. There are other QA datasets for which researchers exploited external knowledge~\cite{baral2020natural}, but the type of KG that they used is not focused on commonsense.

However, most of the retrieved knowledge is not semantically related to the questions for these models. 
In contrast, our MRRG model designed a KG attention module to extract the most relevant knowledge that reduces noise and builds up a relevant subgraph with high quality. 
We will describe more details about the differences between baseline models and MRRG in next sections. 

\begin{figure}
\centering
\includegraphics[width=0.45\textwidth,height=55]{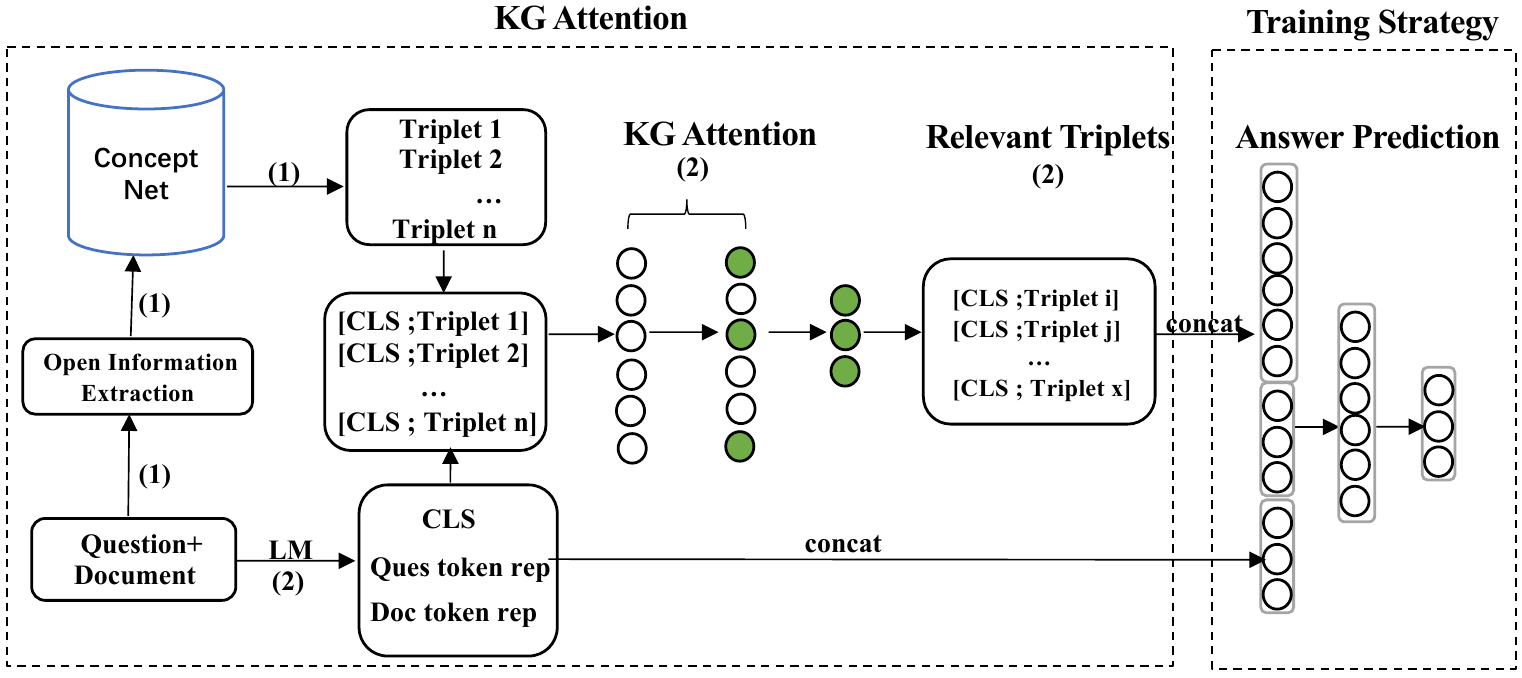}
\caption{The architecture of training the subgraph construction module.}
\label{fig:pretrain}
\end{figure}

\bibliography{acl2022}
\bibliographystyle{acl_natbib}